\documentclass{article}

\PassOptionsToPackage{numbers, compress, sort}{natbib}
\usepackage[final]{neurips_2019}

\usepackage{booktabs}
\usepackage{xspace}
\usepackage{amsfonts}

\usepackage{subcaption}
\usepackage{multirow}
\usepackage{enumitem}
\usepackage{hyperref}
\usepackage{url}
\usepackage{color}
\usepackage[pdftex]{graphicx}
\usepackage{longtable}

\newcommand{\para}[1]{\noindent {\bf #1}\xspace}
\newcommand{\indentpara}[1]{{\bf #1}\xspace}
\newcommand{\figref}[1]{Fig.~\ref{#1}\xspace}
\newcommand{\secref}[1]{\S\ref{#1}\xspace}
\newcommand{\personalsurvey}{{\em personal} survey\xspace}
\newcommand{\expertsurvey}{{\em expert} survey\xspace}
\newcommand{\bothsurveys}{{\em personal} and {\em expert} surveys\xspace}

\newif{\ifhidecomments}
\hidecommentsfalse
\ifhidecomments
    \newcommand{\chenhao}[1]{}
    \newcommand{\brian}[1]{}
\else
    \newcommand{\chenhao}[1]{\textcolor{blue}{[#1 ---\textsc{ct}]}}
    \newcommand{\brian}[1]{\textcolor{red}{[#1 ---\textsc{bl}]}}
\fi

\title{Ask not what AI can do,
but what AI should do:\\ Towards a framework of task delegability}

\author{\\
  Brian Lubars\\
  University of Colorado Boulder\\
  \texttt{brian.lubars@colorado.edu}\\
  \And\\
  Chenhao Tan\\
  University of Colorado Boulder\\
  \texttt{chenhao.tan@colorado.edu}\\
}

\begin{document}

\maketitle

\begin{abstract}
While artificial intelligence (AI) holds promise for addressing societal challenges, issues of exactly which tasks to automate and 
to what extent to do so remain understudied. 
We approach this problem of task delegability from a human-centered perspective by developing a framework on human perception of task delegation to AI.
We consider four high-level factors that can contribute to a delegation decision: motivation, difficulty, risk, and trust.
To obtain an empirical understanding of human preferences in different tasks, we build a dataset of 100 tasks from academic papers, popular media portrayal of AI, and everyday life, and administer a survey based on our proposed framework. 
We find little preference for full AI control and a strong preference for machine-in-the-loop designs, in which humans play the leading role. 
Among the four factors, trust is the most 
correlated with human preferences of optimal human-machine delegation.
This framework represents a first step towards characterizing human preferences of AI automation across tasks. 
We hope this work 
encourages 
future efforts towards understanding such individual attitudes; 
our goal is to inform the public and the AI research community rather than
dictating any direction in technology development.
\end{abstract}

\section{Introduction}

Recent developments in machine learning have led to significant excitement about the promise of artificial intelligence.
\citet{ng2017artificial} claims that ``artificial intelligence is the new electricity.''
Artificial intelligence indeed approaches or even outperforms human-level intelligence in critical domains such as hiring, medical diagnosis, and judicial systems \cite{cheng2016computer,erel2018selecting,fakoor2013using,kleinberg-bail,litjens2016deep}.
However, we also observe growing concerns about which problems
are appropriate applications of AI.
For instance, a recent study used deep learning to predict sexual orientation from images \cite{wang2018deep}.
This study has caused controversy \cite{nyt:sexual,mashable:sexual}: Glaad and the Human Rights Campaign denounced the study as ``junk science'' that ``threatens the safety and privacy of LGBTQ and non-LGBTQ people alike'' \cite{glaad:sexual}.
In general, researchers also worry about the impact on jobs and the future of employment \cite{Frey,schwab2017fourth,susskind2015future}.

Such excitement and concern begs a fundamental question at the interface of artificial intelligence and human society: which tasks should be delegated to AI, and in what way?
To answer this question, we need to at least consider two dimensions.
The first dimension is capability. 
Machines may excel at some tasks, but struggle at others; this area has been widely explored since Fitts first tackled function allocation in 1951 \cite{fitts-haba-maba,sheridan-parasuraman2000model,price-matrix}.
The goal of AI research has also 
typically focused on pushing the boundaries of machine ability and exploring {\em what AI can do}.

The second dimension is human preferences, i.e., what role humans would like AI to play.
The automation of some tasks is celebrated, while others should arguably {\em not} be automated for reasons beyond capability alone.
For instance, automated civil surveillance may be disastrous from ethical, privacy, 
and legal standpoints.
Motivation 
is another reason: no matter
the quality of machine text generation,
it is unlikely that an aspiring writer will derive the same satisfaction or value from delegating their writing to a
fully automated system.
Despite the clear importance of understanding human preferences, the question of {\em what AI should do} remains understudied in 
AI 
research.

In this work, we present the first empirical study to understand how different factors relate to human preferences of {\em delegability}, i.e., to what extent AI should be involved.
Our contributions are threefold.
First, building on prior literature on function allocation, mixed-initiative systems, and trust and reliance on machines
\cite[inter alia]{lee,sheridan-parasuraman2000model,Horvitz-1999},
we develop a framework of four 
factors --- motivation, difficulty, risk, and trust --- 
to explain task delegability. %
Second, we construct a dataset of diverse tasks ranging from those found in academic research to ones that people routinely perform in daily life.
Third, we conduct a survey to solicit human evaluations of the four factors and delegation preferences, and validate the effectiveness of our framework.
We find that our survey participants seldom prefer full automation, but value AI assistance.
Among the four factors, 
trust is the most correlated with human preferences of delegability.
However, the need for interpretability, a component in trust, does not show a strong correlation with delegability.
Our study contributes towards a framework of task delegability and an evolving database of tasks and their associated human preferences.

\section{Related Work}

As machine learning grows further embedded in society, human preferences of AI automation gain relevance. 
We believe surveying, tracking, and understanding such preferences is important.
However, apart from human-machine integration studies on specific tasks,
the primary mode of thinking in AI research is towards automation.
We summarize related work in three main areas.

\para{Task allocation and delegation.} 
Several studies have proposed theories of delegation in the context of general automation \cite{sheridan-parasuraman2000model,price-matrix,castelfranchi-deleg-1998,milewski1}. 
Function allocation examines how to best divide tasks based on human and machine abilities \cite{fitts-haba-maba,sheridan-parasuraman2000model,price-matrix}. 
\citet{castelfranchi-deleg-1998} emphasize the role of risk, uncertainty, and trust in delegation, which we build on in developing our framework.
\citet{milewski1} suggest that people may not want to delegate to machines in tasks characterized by low trust or low confidence, where automation is unnecessary, or where automation does not add to utility.
In the context of jobs and their susceptibility to automation, \citet{Frey} find social, creative, and perception-manipulation requirements to be good prediction criteria for machine ability.
\citeauthor{sheridan-parasuraman2000model}'s {\em Levels of Automation} is the closest to our work \cite{sheridan-parasuraman2000model}. However, their work is primarily concerned with performance-based criteria (e.g., capability, reliability, cost), while our 
interest involves human preferences.

\para{Shared-control design paradigms.} Many tasks are amenable to a mix of human and machine control.
Mixed-initiative systems and collaborative control have gained traction over function allocation,
driven by the need for flexibility and adaptability, and the importance of a user's goals in optimizing value-added automation
\cite{ieee-hri-teamwork,Horvitz-1999,johnson2014coactive}.

We broadly split work on shared-control systems into two categories.
We find this split more flexible and practical for our application than the {\em Levels of Automation}.
On one side, we have human-in-the-loop machine learning designs, wherein humans assist machines. 
The human role is to oversee, handle edge cases, augment training sets, and refine system outputs.
Such designs enjoy prevalence in applications from vision and recognition to translation \cite{branson2010visual,fails2003interactive,Green-NLT-2015}.
Alternatively, a machine-in-the-loop paradigm places the human in the primary position of action and control while the machine assists.
Examples of this include a creative-writing assistance system that generates contextual suggestions \cite{clark2018creative,Roemmele-writing}, 
and predicting situations in which people are likely to make judgmental errors in decision-making \cite{AndersonKM16}. 
Even tasks which should not be automated {\em may} still benefit from machine assistance \citep{green2019disparate,green2019principles,lai+tan:19}, especially if human performance is not the upper bound as \citeauthor{kleinberg-bail} found in judge bail decisions \cite{kleinberg-bail}.

\para{Trust and reliance on machines.} 
Finally, we consider the community's interest in trust.
As automation grows in complexity, complete understanding becomes impossible; trust serves as a proxy for rational decisions in the face of uncertainty, and appropriate use of technology becomes 
critical~\cite{lee}. 
As such, calibration of trust continues to be a popular avenue of research \cite{lewandowsky,gombolay2016robotic}. 
\citet{lee} identify three bases of trust in automation: performance, process, and purpose. 
{\em Performance} describes the automation's ability to reliably achieve the operator's goals. 
{\em Process} describes the inner workings of the automation; examples include dependability, integrity, and interpretability (in particular, interpretable ML has received significant interest \cite{ribeiro2018anchors,ribeiro2016should,kim2016examples,doshi2017towards}).
Finally, {\em purpose} refers to the intent behind the automation and its alignment with the user's goals. 

\section{A Framework for Task Delegability}

\begin{figure}
\centering
\begin{minipage}{0.62\textwidth}
\small
\centering
\captionsetup{type=table} %
    \begin{tabular}{p{1.5cm}p{6.3cm}}
        \toprule
        \textbf{Factors} & \textbf{Components} \\
        \midrule
        Motivation & Intrinsic motivation, goals, utility\\
        \midrule
        Difficulty & Social skills, creativity, effort required, expertise required, human ability \\
        \midrule
        Risk & Accountability, uncertainty, impact \\
        \midrule
        Trust & Machine ability, interpretability, value alignment \\        
        \bottomrule
    \end{tabular}
    \caption{An overview of the four factors in our AI task delegability framework. 
    }
    \label{table:framework}
\end{minipage}
\hfill
\begin{minipage}{0.35\textwidth}
\centering
  \includegraphics[width=\textwidth]{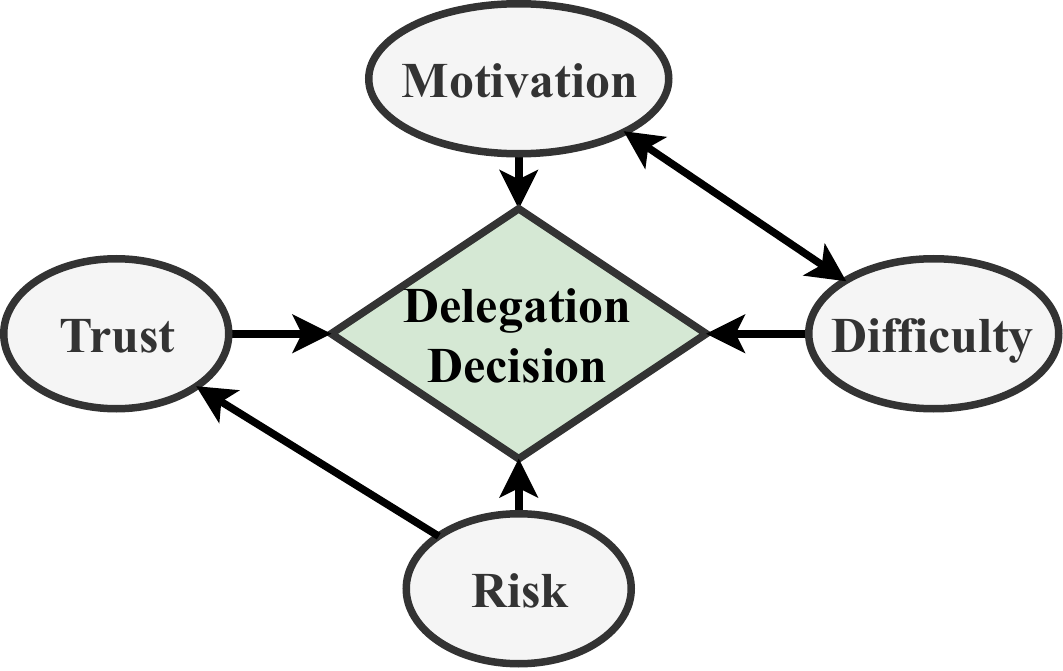}
  \caption{Factors behind task delegability.}
  \label{fig:framework}
\end{minipage}
\end{figure}

To explain 
human preferences of task delegation to AI, we develop a framework with four factors: a person's \textbf{motivation} in undertaking the task, their perception of the task's \textbf{difficulty}, their perception of the \textbf{risk} associated with accomplishing the task, and finally their \textbf{trust} in the AI agent.
We choose these factors because motivation, difficulty, and risk
respectively cover why a person chooses to perform a task, the process of
performing a task, and the outcome, while trust captures the
interaction between the person and the AI. 
We now explain the four factors, situate them in literature, and present the statements used in the surveys to capture each component.
Table \ref{table:framework} presents an overview. %

\indentpara{Motivation.}
Motivation is an energizing factor that helps initiate, sustain, and regulate task-related actions
by directing our attention towards goals or values \cite{Locke-motivation-2000,latham2005work}.
Affective (emotional) and cognitive processes are thought to be collectively responsible for driving action, so we consider \textbf{\em intrinsic motivation} and \textbf{\em goals} as two components in motivation \cite{Locke-motivation-2000}.
We also distinguish between learning goals and performance goals, as indicated by Goal Setting Theory \cite{locke-goal}. 
Finally, the expected \textbf{\em utility} of a task captures its value from a rational cost-benefit analysis perspective \cite{Horvitz-1999}.
Note that a task may be of high intrinsic motivation yet low utility, e.g., reading a novel.
Specifically, we use the following statements to measure these motivation components in our surveys:

\begin{enumerate}[topsep=-4pt,itemsep=-2pt,leftmargin=*]
\item \textbf{Intrinsic Motivation:} ``I would feel motivated to perform this task, even without needing to; for example, it is fun, interesting, or meaningful to me.''
\item \textbf{Goals:} ``I am interested in learning how to master this task, not just in completing the task.''
\item \textbf{Utility:} ``I consider this task especially valuable or important; I would feel committed to completing this task because of the value it adds to my life or the lives of others.''
\end{enumerate}

\indentpara{Difficulty.} Difficulty is a subjective measure reflecting the cost of performing a task. For delegation, we frame difficulty as the interplay between task requirements and the ability of a person to meet those requirements.
Some tasks are difficult because they are time-consuming or laborious; others, because of the required training or expertise. To differentiate the two, we include \textbf{\em effort required} and \textbf{\em expertise required} as components in difficulty. The third component, \textbf{\em belief about abilities possessed}, can also be thought of as task-specific self-confidence (also called self-efficacy \cite{Bandura-89}) and has been empirically shown to predict allocation strategies between people and automation \cite{Lee-1994}. 
Additionally, we contextualize our difficulty measures with two specific skill requirements: the amount of \textbf{\em creativity} and \textbf{\em social skills} required. We choose these because they are considered more difficult for machines than for humans \cite{Frey}.

\begin{enumerate}[topsep=-4pt,itemsep=-2pt,leftmargin=*]
\item \textbf{Social skills:} ``This task requires social skills to complete.''
\item \textbf{Creativity:} ``This task requires creativity to complete.''
\item \textbf{Effort:} ``This task requires a great deal of time or effort to complete.''
\item \textbf{Expertise:} ``It takes significant training or expertise to be qualified for this task.''
\item \textbf{(Perceived) human ability:} ``I am confident in [my own/a qualified person's] ability to complete this task.''
  \footnote{We flip this component %
  in coherence analysis (\figref{fig:component_correlations}) so that higher lack of \textit{confidence in human ability}
indicates higher difficulty.}
\end{enumerate}

\indentpara{Risk.} Real-world tasks involve uncertainty and risk in accomplishing 
the task, so a rational decision on delegation involves more than just cost and benefit. 
Delegation amounts to a bet: a choice considering the probabilities of accomplishing the goal against the risks and costs of each agent \cite{castelfranchi-deleg-1998}. 
\citet{perkins-risk-2010} define risk practically as a ``probability of harm or loss,''
finding that people rely on automation less as the probability of mortality increases.
Responsibility or accountability may play a role if delegation is seen as a way to share blame \cite{lewandowsky,accountability}. 
We thus decompose risk into 
three components:
personal \textbf{\em accountability} for the task outcome; the \textbf{\em uncertainty}, or the probability of errors; and the scope of \textbf{\em impact}, or cost or magnitude of those errors.

\begin{enumerate}[topsep=-4pt,itemsep=-2pt,leftmargin=*]
\item \textbf{Accountability:} ``In the case of mistakes or failure on this task, someone needs to be held accountable.''
\item \textbf{Uncertainty:} ``A complex or unpredictable environment/situation is likely to cause this task to fail.''
\item \textbf{Impact:} ``Failure would result in a substantial negative impact on my life or the lives of others.''
\end{enumerate}

\indentpara{Trust.} 
Trust captures how people deal with risk or uncertainty.
We use 
the definition of trust as ``the attitude that an agent will help achieve an individual's goals in a situation characterized by uncertainty and vulnerability'' \cite{lee}.
Trust is generally regarded as the most salient factor in reliance on automation \cite{lee,lewandowsky}.
Here, we operationalize trust as a combination of perceived \textbf{\em ability} of the AI agent, agent \textbf{\em interpretability} (ability to explain itself), and perceived \textbf{\em value alignment}. 
Each of these corresponds to a component of trust 
in \citet{lee}: performance, process, and purpose.
\begin{enumerate}[topsep=-4pt,itemsep=-2pt,leftmargin=*]
  \item \textbf{(Perceived) machine ability:} ``I trust the AI agent's ability to reliably complete the task.''
  \item \textbf{Interpretability:} ``Understanding the reasons behind the AI agent's actions is important for me to trust the AI agent on this task (e.g., explanations are necessary).''
  \footnote{We flip this component 
  in coherence analysis (\figref{fig:component_correlations} and \ref{fig:trust-corr})
    so that higher lack of \textit{need for interpretability} indicates higher trust.}
  \item \textbf{Value alignment:} ``I trust the AI agent's actions to protect my interests and align with my values for this task.''
\end{enumerate}

\indentpara{Degree of delegation.} We develop this framework of motivation, difficulty, risk, and trust to explain human preferences of delegation.
To measure human preferences,
we split the degree of delegation into the following four categories (refer to Supplementary Material for the wordings in the survey):

\begin{enumerate}[topsep=-4pt,itemsep=-2pt,leftmargin=*]
  \item \textbf{No AI assistance:} the person does the task completely on their own (henceforth, ``human only'').
\item \textbf{The human leads and the AI assists:} the person does the task mostly on their own, but the AI offers recommendations or help when appropriate (e.g., human gets stuck or AI sees possible mistakes) (henceforth, ``machine in the loop''). 
  \item \textbf{The AI leads and the human assists:} the AI performs the task, but asks the person for suggestions/confirmation when appropriate (henceforth, ``human in the loop''). 
\item \textbf{Full AI automation}: decisions and actions are made automatically by the AI once the task is assigned; no human involvement (henceforth, ``AI only'').
\end{enumerate}

\figref{fig:framework} presents our expectation of how 
motivation, difficulty, risk, and trust 
relate to delegability.
Motivation describes how invested someone is in the task, including how much effort they are willing to expend,
while difficulty determines the amount of effort the task requires. 
Therefore, we expect difficulty and motivation to relate to each other: we hypothesize that people are more likely to delegate tasks which they find difficult (or have low confidence in their abilities), and less likely to delegate tasks which they are highly invested in. 
Risk reflects uncertainty and vulnerability in performing a task, the situational conditions necessary for trust to be salient \cite{lee}.
We thus expect risk to moderate trust. 
Finally, we hypothesize that the correlation between components within each factor should greater than that across different factors,
i.e., factors should show coherence in component measurements.

\section{A Task Dataset and Survey Results}

To evaluate our framework empirically, we build a database of diverse tasks covering settings ranging from academic research to daily life, and develop and administer a survey to gather perceptions of those tasks under our framework. 
We examine survey responses through both quantitative analyses and qualitative case studies.

\subsection{A Dataset of Tasks}
We choose 100 tasks drawn from academic conferences, popular discussions in the media, well-known occupations, and mundane tasks people encounter in their everyday lives. 
These tasks are generally relevant to current AI research and discussion, and present challenging delegability decisions with which to evaluate our framework.
Example 
tasks can be found in 
\secref{sec:casestudy}.
To additionally balance the variety of tasks chosen, we categorize them as art, creative, business, civic, entertainment, health, living, and social, and keep a minimum of 7 tasks of each (a task can belong to multiple categories; refer to Supplementary Material for details).
Ideally, the 
tasks would cover the entire automation ``task space''; our task set is intended as a reasonable starting point.

Since some tasks, e.g., medical diagnosis, require expertise, and since motivation does not apply if the subject does not personally perform the task, we develop two survey versions.

\begin{itemize}[topsep=-5pt,itemsep=-2pt,leftmargin=*]
    \item {\bf Personal survey.} We include all the four factors in Table~\ref{table:framework} and ask participants ``If you were to do the given (above) task, what level of AI/machine assistance would you prefer?''
    \item {\bf Expert survey.} We include only difficulty, risk, and trust, and ask participants ``If you were to ask someone to complete the given (above) task, what level of AI/machine assistance would you prefer?''
\end{itemize}

Following a general explanation, our survey begins by asking subjects for basic demographic information.
Subjects are then presented one randomly-selected task from our set. They evaluate the task under each component in our framework (see Table \ref{table:framework}) according to a five-point Likert scale.
Finally, subjects select 
the degree of 
delegation they would prefer for the task
from the following four choices: Full Automation, AI leads and human assists, Human leads and AI assists, or No AI assistance.
Note that subjects are not told which factor each question measures beyond the question text itself, and can choose the degree of delegation independently of our framework.

We administer this 5-minute survey on 
Mechanical Turk. 
To improve the quality of surveys, we require that participants have completed 200 HITs with at least a 99\% acceptance rate and are from the United States.
We additionally add two attention check questions to ensure participants read the survey carefully.
Subjects are paid \$0.80 upon completing the survey and passing the checks; otherwise the data is discarded.
We record 1000 survey responses: 500 each for the personal and the expert versions, composed of 5 responses for each of the 100 tasks.
Finally, we further discard responses that are identical in each component (e.g., marking ``Agree'' for all components), resulting in 495 and 497 responses for the personal and expert versions, respectively. This leaves 8 tasks with 4 responses rather than 5.
We obtain a gender-balanced sample with 525 males, 463 females, and 4 identifying otherwise.
The dataset is available at \url{http://delegability.github.io}.

\subsection{Survey Results}

In this section, we begin by examining the overall distribution of the delegability preferences, then investigate the relation between components in our framework and the delegability labels.

\begin{figure}
\footnotesize
\centering
\begin{minipage}{\textwidth}
\centering
\begin{subfigure}[t]{0.23\textwidth}
  \centering
  \includegraphics[width=\textwidth]{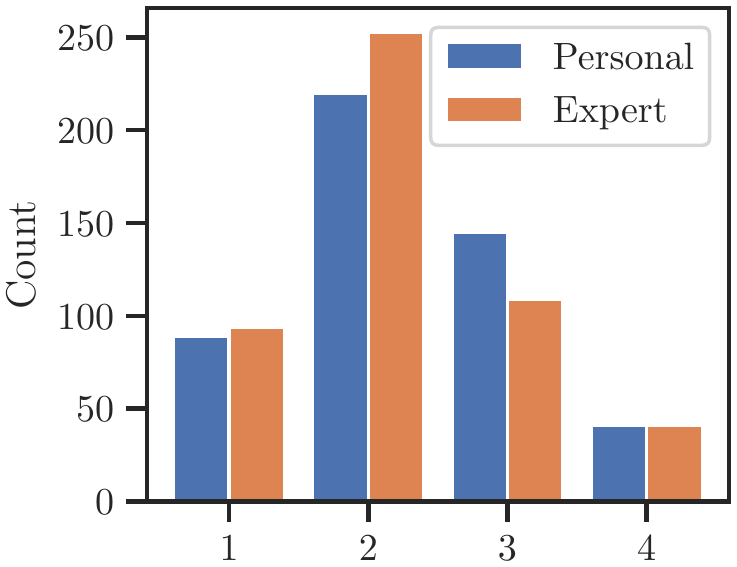}
  \caption{Distribution of individual preferences.}
  \label{fig:individual_label_dist}
\end{subfigure}
\hfill
\begin{subfigure}[t]{.23\textwidth}
  \centering
  \includegraphics[width=\textwidth]{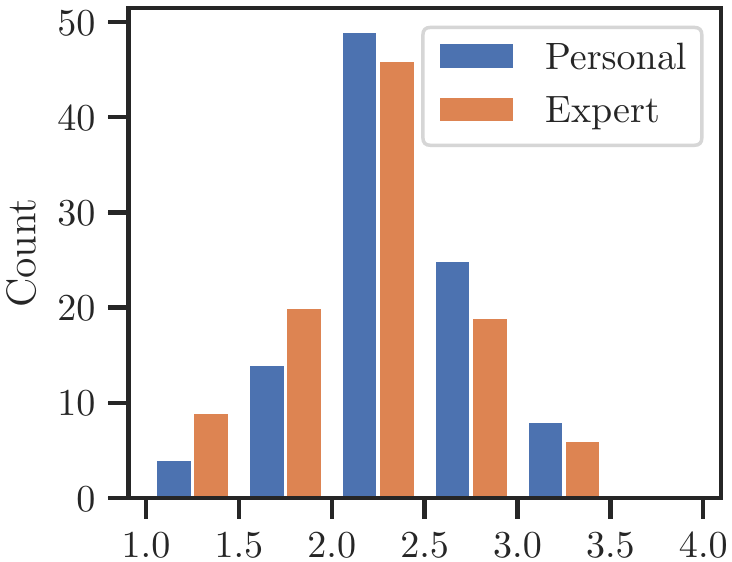}
  \caption{Dist. of average preferences for each task.}
  \label{fig:average_label_dist}
\end{subfigure}
\hfill
\begin{subfigure}[t]{0.23\textwidth}
        \centering
\includegraphics[width=\textwidth]{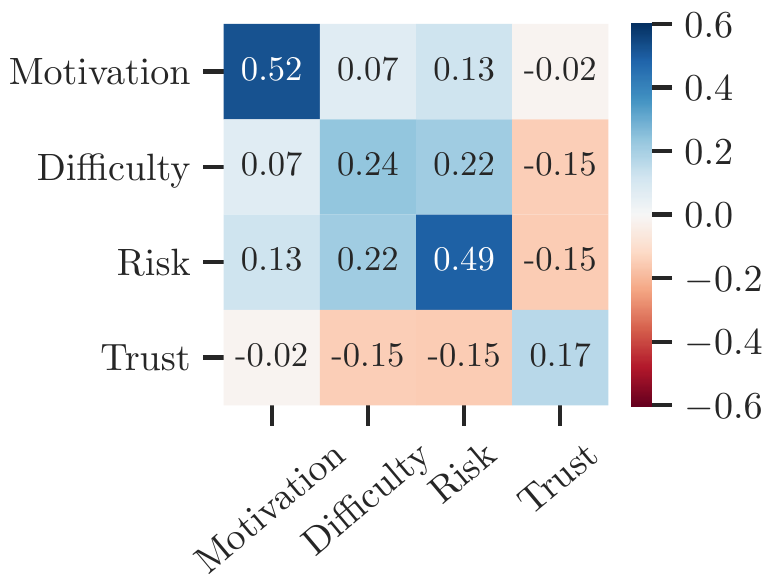}
  \caption{Average pairwise correlation between components in the four factors.}%
\label{fig:component_correlations}
\end{subfigure}
    \hfill
\begin{subfigure}[t]{0.23\textwidth}
        \centering
        \includegraphics[width=\textwidth]{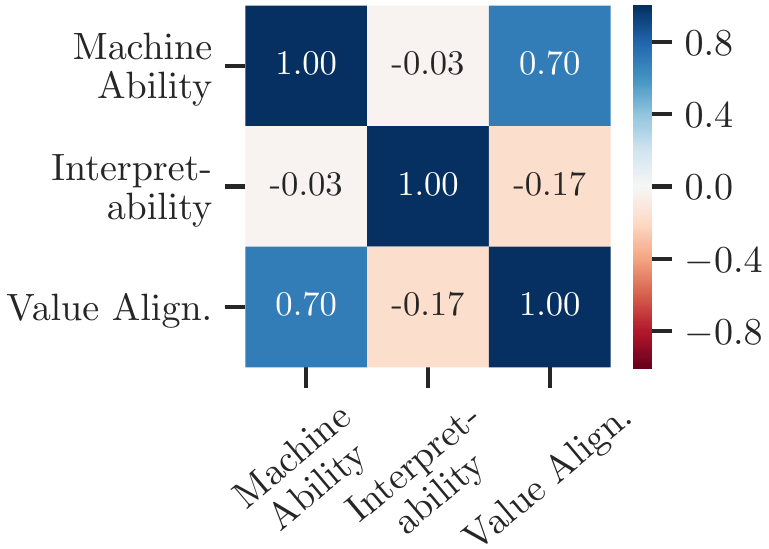}
        \caption{Correlation of components in the trust factor.}
        \label{fig:trust-corr}
\end{subfigure}
\caption{\figref{fig:individual_label_dist} and \ref{fig:average_label_dist} show that full automation is rarely preferred in our surveys. \figref{fig:component_correlations} and \ref{fig:trust-corr} examine correlations between components. We observe low coherence in trust and difficulty. In particular, interpretability seems distinct from the other two trust components.}
\label{fig:diff-trust-corr}
\end{minipage}
\hfill
\begin{minipage}{\textwidth}
\scriptsize
\captionsetup{type=table} %
    \centering
    \begin{tabular}{@{}llll@{}}
        \toprule
        \textbf{Factor} & \textbf{Component} & \textbf{Personal} & \textbf{Expert} \\
        \midrule
        Motivation     & Utility          & -0.126 ($\dagger$)  & N/A     \\
        Motivation     & Intrinsic motivation  & -0.104 ($\dagger$)   & N/A     \\
        Motivation     & Goals    & NS           & N/A     \\
        \midrule
        Difficulty     & Social skills   & -0.303 (***)  & -0.294 (***) \\
        Difficulty     & Creativity & -0.223 (***)  & -0.290 (***) \\
        Difficulty     & Human ability   & NS            & -0.160 (**) \\
        Difficulty     & Effort required      & NS            & -0.124 ($\dagger$)  \\
        Difficulty     & Expertise required   & NS            & -0.120 ($\dagger$)  \\
        \midrule
        Risk           & Uncertainty     & NS            & -0.135 ($\dagger$)  \\
        Risk           & Accountability  & NS            & -0.123 ($\dagger$)  \\
        Risk           & Impact          & NS            & -0.106 ($\dagger$)   \\
        \midrule
        Trust          & Machine ability & 0.520 (***)   &  0.593 (***) \\
        Trust          & Value alignment & 0.481 (***)   &  0.522 (***) \\
        Trust          & Process         & NS            & NS      \\
        \bottomrule
    \end{tabular}
    \caption{Pearson correlation of framework components with the delegabilty label for individual responses to the personal and expert surveys.
    $p$-values were computed by aggregating over individual Likert ratings separately for the \bothsurveys, resulting in $25$ 
    tests in total.
    Significance after Bonferroni correction is indicated by 
    *** for $p < 0.001$, ** for $p < 0.01$, * for $p < 0.05$, and NS for $p >= 0.05$.
    Results that were $p < 0.05$ prior to correction are indicated by $\dagger$.
    }
    \label{table:correlations}
\end{minipage}
\end{figure}

\para{Participants seldom choose ``AI only'' and prefer designs where humans play the leading role.}
Participants labeled the delegability of tasks 
ranging from 1 (``Human only'') to 4 (``AI only'').
\figref{fig:individual_label_dist} presents the distribution.
In both the \bothsurveys,
4 (``AI only'') is seldom chosen.
Instead, both distributions peak at 2, indicating strong preferences towards machine-in-the-loop designs.
This result becomes even more striking when averaging the five responses received per task, concentrating almost half the mass 
between 2 and 2.5 --- again indicating a preference for machine-in-the-loop designs.
In fact, after averaging responses, we find that none of the 100 tasks yield an overall preference for full automation ($\geq$~3.5).
Taken together, these results imply that people prefer 
humans to keep control over the vast majority of 
tasks, yet are also open to AI assistance.

If we view our surveys as a labeling exercise,
the agreement 
between individuals
is low but is relatively higher in the \expertsurvey than the \personalsurvey:
the Krippendorff's $\alpha$ is 0.063 in the \personalsurvey and is 0.174 in the \expertsurvey \cite{hayes2007answering}.
The lower agreement in the \personalsurvey is consistent with heterogeneity between individuals.
Two of the most contentious \personalsurvey tasks were: ``Cleaning your house'' and ``Breaking up with your romantic partner''.

\para{Trust is most correlated with human preferences of automation.}
Consistent with our hypothesis in \figref{fig:framework}, trust is generally positively correlated with delegability.
Table \ref{table:correlations} shows the 
component correlations with the delegability labels.
After Bonferroni correction, 
5 out of the 11 components are significantly correlated with the delegability label
in the \expertsurvey, 
while only 4 of the 14 components are in the \personalsurvey.
Trust takes the top two spots in both.
Difficulty, the only other significantly correlated factor after trust, is generally negatively correlated with delegability.
In particular, the creative and social skill requirements are the next highest correlations in both surveys,
suggesting specific skills may form a stronger basis for delegation to AI than more subjective difficulty measures (e.g., self-confidence).

Next, we highlight three exceptions:
1) 
Contrary to our hypothesis,
we did not find statistically significant correlations between the risk or motivation factors and delegability after Bonferroni correction.
2) The {\em interpretability (process)} component of trust is not correlated with delegability.
3) Confidence in human ability (within the difficulty factor; lower confidence indicates higher difficulty) is not correlated with delegability in the \personalsurvey, but is actually negatively correlated in the \expertsurvey (the lower the confidence, the more delegable).
This differs from the general trend of rating more difficult tasks as less delegable,
suggesting that {\em people 
prefer experts to accept AI assistance on low-confidence tasks, but perhaps are less willing to do so themselves}.

\para{Factors are generally ``coherent'', but risk components
have only weak correlation with trust components.}
We next study the correlation between components in our framework.
We focus on the \personalsurvey here because it has all four factors, but results are consistent in the \expertsurvey.
\figref{fig:component_correlations} presents the average pairwise component correlations between the four factors: the correlation along the diagonal is generally higher than the off-diagonal ones.
This finding confirms that factors are generally ``coherent'', affirming our categorization of the components within them.

Comparing the factor relations to our expectation in \figref{fig:framework} (see the correlation graph in the Supplementary Material for detailed relations between components), we find that motivation and difficulty are indeed correlated, most notably between self-confidence and intrinsic motivation in the \personalsurvey ($\rho = 0.41$, people enjoy doing what they are good at).
However, contrary to our expectation, 
components in risk have only weak connections with trust, while difficulty is correlated with trust (through the social and creative skill requirements) and to risk in \bothsurveys.

Coherence is lower in difficulty and trust than in risk and motivation.
To investigate this, 
we zoom into the correlation matrix in \figref{fig:trust-corr} to show individual components of trust.
We observe interpretability has little correlation with machine ability and negative correlation with value alignment, suggesting that the need for explanation is independent of whether the machine is perceived as capable, and perhaps higher when machine is perceived as benign. 
In comparison, interpretability is most strongly correlated with risk. In the \personalsurvey, it is also connected to motivation through utility and learning goals.
Thus risky and important tasks,
which people want to learn,
tend to require more interpretability; 
but this may not be tied to their willingness to delegate.

\begin{minipage}{\textwidth}
  \scriptsize
  \captionsetup{type=table}
  \centering
  \begin{tabular}{p{4.2cm}@{\hskip 0.5em}l*{7}{@{\hskip 0.5em}l}}
    \toprule
    \textbf{Task Description} & \textbf{Social}     & \textbf{Expertise} & \textbf{Human}       & \textbf{Account-}    & \textbf{Impact} & \textbf{Machine}     & \textbf{Interpret-} & \textbf{Delegability} \\
                             & \textbf{Skills (D)} & \textbf{Req (D)}   & \textbf{Ability (D)} & \textbf{ability (R)} & \textbf{(R)}    & \textbf{Ability (T)} & \textbf{ability (T)}              & \\ 
    \midrule
    Medical diagnosis: flu            & 3.4  & 4.2  & 4.6  & 4    & 4.2  & 3    & 4.2 & 2.4 \\
    Medical treatment planning: flu   & 3.6  & 3.6  & 4.4  & 4    & 3.4  & 3.6  & 3.8 & 2.4 \\
    Explaining diagnosis \& treatment 
    options to a patient: flu         & 3.4  & 3.6  & 4.2  & 3.8  & 3.2  & 3.8  & 3.2 & 2.2 \\ \midrule
    Medical diagnosis: depression     & 4.4  & 4.6  & 4.6  & 4.2  & 3.8  & 2.8  & 3.4 & 2.2 \\
    Medical treatment planning: 
    depression                        & 3.8  & 3.4  & 4.4  & 4.2  & 4.4  & 3    & 4   & 2   \\
    Explaining diagnosis \& treatment 
    options to a patient: depression  & 4.4  & 4.4  & 4.2  & 4.4  & 4.4  & 2.2  & 4.4 & 1.6 \\ \midrule
    Medical diagnosis: cancer         & 2.6  & 5    & 3.6  & 3.8  & 4.8  & 2.4  & 3.4 & 2   \\
    Medical treatment planning: 
    cancer                            & 3.6  & 4.6  & 4.8  & 4.4  & 4.8  & 2.4  & 3.8 & 1.6 \\
    Explaining diagnosis \& treatment 
    options to a patient: cancer      & 4.4  & 4.4  & 4.2  & 4.2  & 4.6  & 2.4  & 2.6 & \textcolor{red}{1.4} \\ \midrule
    New employee hiring decisions     & 4.4  & 3.6  & 3.8  & 3.8  & 3.8  & 2.4  & 4.4 & 2.2 \\
    Judging a defendant's 
    recidivism risk                   & 3.8  & 4    & 4.4  & 4.4  & 4.6  & 2.4  & 3.6 & 1.8 \\ \bottomrule
  \end{tabular}
  \caption{A case study of tasks from the \expertsurvey. See full names of tasks in the supplementary material. Note that we do not flip any component in these case study tables. }
    \label{table:case-expert}
\end{minipage}

\begin{minipage}{\textwidth}
  \scriptsize
  \captionsetup{type=table}
  \centering
  \begin{tabular}{p{4.2cm}@{\hskip 0.5em}l*{7}{@{\hskip 0.5em}l}}
    \toprule
    \textbf{Task Description} & \textbf{Social}     & \textbf{Expertise} & \textbf{Human}       & \textbf{Account-}    & \textbf{Impact} & \textbf{Intrinsic} & \textbf{Machine}     & \textbf{Delegability} \\
                     & \textbf{Skills (D)} & \textbf{Req (D)}   & \textbf{Ability (D)} & \textbf{ability (R)} & \textbf{(R)}    & \textbf{(M)} & \textbf{Ability (T)} \\ \midrule
    Serving on jury duty  & 4.6  & 2.4  & 4.6  & 4.6  & 4.2  & 3.8  & 1.2  & \textcolor{red}{1.4} \\
    New employee hiring decisions & 4  & 3.6  & 3.6  & 4.2  & 3  & 2.6  & 1.8  & 1.8 \\
    Reading bedtime stories to your child  & 4  & 2  & 4.4  & 2.6  & 1.8  & 4.8  & 3.2  & 1.8 \\
    Scheduling an important business meeting & 3.6  & 2  & 4.4  & 3.2  & 3  & 2.6  & 3.6  & 3 \\ \bottomrule
  \end{tabular}
  \caption{A case study of tasks from the \personalsurvey. Refer to \url{https://delegability.github.io/table.html} for live demo.}
    \label{table:case-personal}
  \vspace{-0.1in}
\end{minipage}

\subsection{Case Studies}
\label{sec:casestudy}

To further illustrate the operation of our delegability framework, we present averaged responses to selected tasks from the \expertsurvey in Table \ref{table:case-expert} and the \personalsurvey in Table \ref{table:case-personal}.
For the expert case studies, we examine responses to medical diagnosis, recidivism risk, and hiring.
Next, we observe the effects of motivation on some personal tasks.
Finally, some tasks such as hiring are suitable for both experts and laypeople, allowing us to compare the personal and expert contexts.
Though clearly important, we observe trust alone cannot explain differences in delegability preferences.

\para{Expert Survey Case Studies.}
The medical domain is often considered a promising area for AI, but how open are patients to delegating different aspects of their healthcare to AI?
We compare three medical tasks (diagnosis, treatment, and explanation), each with three different illness contexts (the flu, depression, and cancer).
Intuitively, flu-related tasks are seen as the most delegable -- even approaching a delegability of 3 (human-in-the-loop) -- while cancer is the least.
Correspondingly, the flu-related tasks are perceived as less difficult (lower social skill and expertise requirements, higher confidence in the human expert), less risky (lower impact and accountability), and with higher trust in machine ability.
However, all except cancer explanation are nearest a delegability level of 2 (machine-in-the-loop):
though human control is preferred,
machine assistance is 
valued. %

Fairness decision problems such as recidivism risk and employee hiring decisions are typically characterized as requiring high transparency and accountability. 
From the responses, we see that these tasks are both 
rated as difficult, risky, and with low degrees of trust in AI; in fact, they look similar to the medical tasks under our framework.
Accordingly, we again observe both tasks result in an average preference for machine-in-loop designs.

\para{Personal Survey Case Studies.}
To observe the role of motivation, 
we consider 
``Reading bedtime stories to your child'', a machine-in-the-loop task, and 
``Scheduling an important business meeting with several co-workers'', a human-in-the-loop task.
The former is higher motivation, yet the latter is higher risk. 
The tasks are otherwise similar.
Here, motivation appears to be the deciding factor:
the former's higher motivation makes it less delegable despite the lower risk.

Finally, we compare some personal-context tasks which are similar to the above expert-context case studies.
``Serving on jury duty: deciding if a defendant is innocent or guilty'' is comparable to the high-risk low-trust medical and recidivism tasks, and is also similarly rated one of the least delegable tasks to AI.
For the employee hiring task, respondents rated their self-confidence as near their confidence in an expert. 
Upon directly comparing the other components more closely, we observe 
higher accountability, lower trust, and slightly lower delegability levels in the personal evaluations.

\section{Concluding Discussion}
\vspace{-0.1in}

In this work, we present first steps towards understanding 
human preferences 
in human-AI task delegation decisions.
We develop an intuitive framework of motivation, difficulty, risk, and trust, which we hope will serve as a starting point for reasoning about delegation preferences across tasks. 
We develop a survey to quantify human preferences, and validate the promise of such an empirical approach through correlation analysis and case studies. 
Our findings show a preference for machine-in-the-loop designs and a disinclination towards ``AI Only''.

In developing this framework, our intent is not to suppress 
technology development,
but rather to provide an avenue for more effective human-centered automation.
Human preferences regarding the extent of autonomy, and the reasons and motivations behind these preferences, are an 
understudied yet valuable source of information. 
There is a clear gap in methodologies for both understanding and contextualizing preferences across tasks; it is this gap that we wish to address.

\para{Implications.} 
First,
our finding of trust as the most salient factor behind delegation preferences supports the community's widespread interest in trust and reliance.
We find negative correlations between trust in machine abilities and the social and creative skill requirements. 
These are skills commonly considered difficult for machines, hinting 
that directly measuring task characteristics instead of human assessments of trust may also be an effective approach.
We also note the {\em low} correlation between desired interpretability and delegability to AI, demonstrating the 
complex and intricate relation between interpretable machine learning and trust.

Moreover, our findings show that people do not prefer ``AI only'', instead opting for machine-in-the-loop designs. 
Interestingly, even for low-trust tasks such as cancer diagnosis or babysitting, people are still receptive to the idea of a machine-in-the-loop assistant.
We should explore paradigms that let people maintain high-level control over tasks while leveraging machines as support, as in recent work on clinical decision systems \citep{yang2019unremarkable}.

\para{Limitations.} 
Our framework does not fully explain delegability preferences: the highest measured correlation is $0.59$,
and due to limited data, we do not explore higher-order feature interactions.
Additionally, human preferences are dynamic and survey results 
likely evolve over time. 
Nevertheless, mapping current perceptions enables tracking any future changes, providing a
mechanism to understand how basic changes in factors like machine ability manifest through trust and reliance.

Our exploratory survey methodology also has several limitations.
We abstracted delegability decisions and measured a limited number of factors, thus potentially overstating the importance of trust and overlooking others that occur in real situations.
For example, we 
did not consider situational details like the trust in specific companies, or actual human or machine performance.
Second, we designed the survey to avoid biasing or requiring respondents to understand our framework conceptually.
Training may improve subject calibration and agreement.
We also recommend consideration of individual baseline attitudes towards automation.
Since each participant only filled out one survey, we were unable to determine if individuals were consistently biased 
towards AI, and what effect this might have had on our measurements.
In addition, we chose only four delegability categories, ranging from ``human only'' to ``AI only''. This was a deliberate abstraction choice to handle the wide variety of tasks presented. 
However, since most
responses fell into one of the two shared-control categories, 
future studies may benefit from more fine-grained choices on shared control.

Finally, our empirical survey is based on participants on 
Mechanical Turk.
We use a strict filter and attention checks to guarantee quality, but this sample may not be representative of the general population in the US, or of other populations with different cultural expectations.

\para{Towards a Framework.}
In addition to resolving the above limitations,
we specifically
suggest two characteristics for any framework addressing the 
task delegability question:
1) a characterization of human preferences towards automation and machine control;
and 
2) a characterization of the task space, enabling the generalization of task-specific findings to other domains.

In particular, generalization represents a significant challenge, especially when incorporating human factors.
For instance, it is unclear how to generalize from physicians interacting with AI for cancer diagnosis to judges interacting with AI for recidivism prediction.
While our approach maps tasks to delegability preferences through a common set of task-automation perception factors and 
enables directly comparing tasks with a quantifiable perception distance,
its capability of generalization requires further verification.
Ultimately, we believe an effective quantification of human preferences and task relations will prove invaluable for the community and the public as a whole.

\bibliographystyle{plainnat}
\bibliography{cite}

\newpage
\appendix

\section{Component Correlations}

\begin{figure}[h]
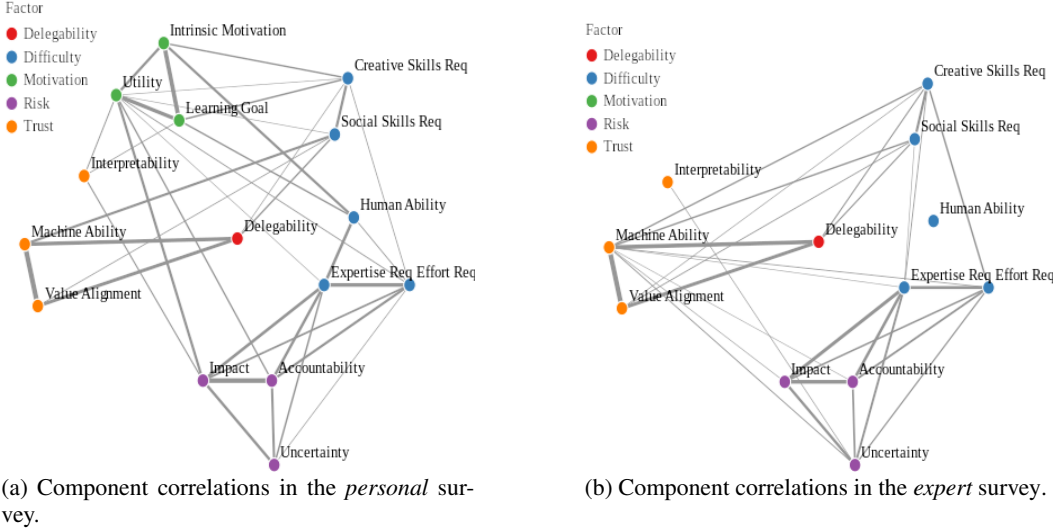

\footnotesize
\centering
\centering
\begin{subfigure}[t]{0.45\textwidth}
  \centering
  \includegraphics[width=\textwidth]{figures/Personal-network-r020.png}
  \caption{Component correlations in the \personalsurvey.}
  \label{fig:personal_network}
\end{subfigure}
\hfill
\begin{subfigure}[t]{.45\textwidth}
  \centering
  \includegraphics[width=\textwidth]{figures/Expert-network-r020.png}
  \caption{Component correlations in the \expertsurvey.}
  \label{fig:expert_network}
\end{subfigure}
  \caption{Component correlations in the \personalsurvey and \expertsurvey. Both figures show only connections with correlation coefficient $|\rho| \geq 0.20$ to prevent overcrowding the graphic.
The weight of the edge is proportional to $|\rho|$.}
\label{fig:survey_networks}
\end{figure}

Due to lack of space, we did not enumerate all correlations between components in the main paper, instead focusing mainly on the correlations between the factors and the task's delegability to AI. 
However, some interesting structure can also be seen in the connections between components themselves.
\figref{fig:survey_networks} shows the strongest component correlations ($|\rho| \geq 0.20$) for direct comparison of our expected framework connections to the observed connections. 

\section{Task Selection \& Methodology}
In selecting our dataset of 100 tasks, our aim is to create a diverse set that is relevant to current AI research and discussion.
Ideally we would compile a large reference set that covers the entire automation ``task space'',
but these 100 tasks are meant as a reasonable starting point.
We source our task set from papers in AI conferences (96 tasks), from occupational descriptions (102 tasks), from media coverage of AI (76 tasks), and from daily life (115 tasks).
For example, ``Analyzing and critiquing aesthetic qualities of photographs or other forms of art'' is drawn from~\citet{chang2017aesthetic}.
The occupational descriptions are adaptations of a subset of \citet{Frey}'s dataset of 702 occupations, which were themselves originally adapted from O*NET. We select a subset which evenly spans the range of predicted occupational susceptibilty to automation \cite{Frey}.

To refine these 389 tasks down to 100 while promoting variety, we then group the tasks into 8 semantic categories: art, creative, business, civic, entertainment, health, living, and social.
A task may belong to multiple categories. For instance, ``Babysitting your child'' is living and social.
Our final set contains a minimum of 7 tasks per category.
The 100 tasks, their sources, and their semantic categories are shown in Table \ref{table:task_table}.
\begin{longtable}{p{0.6\textwidth}p{0.14\textwidth}p{0.15\textwidth}}
    \toprule
    \textbf{Task} & \textbf{Source} & \textbf{Categories}\\ \midrule
Analyzing and critiquing aesthetic qualities of photographs or other forms of art & conference & art, creative \\ \midrule
Choreographing dance moves for a person to perform & conference & art, creative \\ \midrule
Drawing or painting something (making art) & life & art, creative \\ \midrule
Picking a topic to write a short story about & life & art, creative \\ \midrule
Reviewing a book or a movie & life & art, creative \\ \midrule
Writing a blog post & life & art, creative \\ \midrule
Writing a novel or a short story (creative writing) & occupation & art, creative \\ \midrule
Analyzing and sorting legal documents for important information, e.g., to find legal precedents for arguing a case in court (similar to some of what a paralegal might do) & media & business \\ \midrule
Analyzing financial market conditions and executing market orders for a large company (e.g. buy/sell stocks) & conference & business \\ \midrule
Assembling automobiles in a factory & conference & business \\ \midrule
Choosing and ordering food to eat for dinner & life & business \\ \midrule
Coordinating and oversee construction of a building, e.g., consulting with engineers, surveyors, specialists, and construction workers -- similar to some of what an architect might do. & occupation & business \\ \midrule
Deciding which applicants receive a loan from a bank (loan assessment) & conference & business \\ \midrule
Detecting and removing fake/deceptive online reviews (e.g., for hotels or products) & conference & business \\ \midrule
Driving a truck delivering goods/cargo between cities & conference & business \\ \midrule
Driving to work & conference & business \\ \midrule
Establishing compensation/wage/salary level for an employee & occupation & business \\ \midrule
Inferring damage for insurance purposes after a car accident & conference & business \\ \midrule
Interviewing job applicants and rating candidates & media & business \\ \midrule
Monitoring farm animals' (e.g., cows) behavior, predicting health issues, and alerting the farmer. & media & business \\ \midrule
Moving \& packing merchandise in a warehouse for shipping to customers & media & business \\ \midrule
Picking jobs to apply to & conference & business \\ \midrule
Planning menus and developing recipes at a restaurant & occupation & business \\ \midrule
Predicting the sale value of a real estate property & conference & business \\ \midrule
Responding to emails at work & life & business \\ \midrule
Scheduling an important business meeting with several co-workers & life & business \\ \midrule
Serving food to customers at a restaurant & conference & business \\ \midrule
Writing reports and publishing Olympic (or other sports) results, standings, and stats (sports news coverage) & media & business \\ \midrule
Writing reports and publishing updates on House/Senate/gubernatorial races during election day (election news coverage) & media & business \\ \midrule
Cutting, drying, and styling hair, similar to what a barber or hairstylist might do & occupation & business, creative \\ \midrule
Designing new clothing to manufacture and sell (similar to what a fashion designer might do) & conference & business, creative \\ \midrule
Finding products you might be interested in while you're shopping & conference & business, living \\ \midrule
Teaching your child elementary school math (e.g., multiplication, fractions) & life & business, living, civic \\ \midrule
Deciding which applicants to hire as new employees for an open position at work & conference & business, social \\ \midrule
Taking photos of a planned event, such as a wedding or graduation, similar to what a professional photographer might do. & occupation & business, creative, social \\ \midrule
Analyzing and controlling the flow of traffic in a city & conference & civic \\ \midrule
Arguing your case when you're a defendant in a criminal court & life & civic \\ \midrule
Deciding military actions such as whether to launch airstrikes & media & civic \\ \midrule
Detecting/recognizing abnormal or suspicious activities of people in crowds in public places for the purposes of security and safety (similar to part of what a police officer might do) & conference & civic \\ \midrule
Finding and rescuing survivors after earthquakes & conference & civic \\ \midrule
Guiding and explaining exhibits in a museum (similar to what a museum tour guide might do) & conference & civic \\ \midrule
Helping to locate a missing child by searching public spaces & media & civic \\ \midrule
Identifying and flagging fake/deceptive news articles & conference & civic \\ \midrule
Identifying and flagging online hate speech & media & civic \\ \midrule
Identifying people who attended a political rally & media & civic \\ \midrule
In court, determining a defendant's risk (e.g., in committing another crime or missing the court date), to help judges make decisions about bail, sentencing, or parole & media & civic \\ \midrule
Responding to 911-police incident reports, similar to what a patrol officer might do & occupation & civic \\ \midrule
Serving on jury duty: deciding if a defendant is innocent or guilty & life & civic \\ \midrule
Setting tariffs on goods imported from China & media & civic \\ \midrule
Teaching a religion's doctrine and practices to followers, similar to some of the responsibilities of clergy/religious leaders & occupation & civic, social \\ \midrule
Voting in federal elections & life & civic \\ \midrule
Picking a movie to watch & conference & entertainment \\ \midrule
Picking a movie to watch with a group of friends & life & entertainment \\ \midrule
Picking songs to listen to & media & entertainment \\ \midrule
Picking which advertisements to show to people on social media websites & media & entertainment \\ \midrule
Picking which news stories to show to people on social media websites & media & entertainment \\ \midrule
Playing a board game (e.g., monopoly, scrabble) & conference & entertainment \\ \midrule
Playing a competitive game (e.g., dota2, starcraft, poker) & media & entertainment \\ \midrule
Advising people on nutrition/their diet to help improve their health, similar to what a nutritionist might do & occupation & health \\ \midrule
Conducting a risk prognosis assessment for deciding which patients to transfer to the ICU given limited resources (intensive care) & conference & health \\ \midrule
Devising treatment plans for patients sick with the flu & conference & health \\ \midrule
Devising treatment plans for patients with cancer & conference & health \\ \midrule
Devising treatment plans for patients with depression & conference & health \\ \midrule
Diagnosing whether a person has cancer & conference & health \\ \midrule
Diagnosing whether a person has depression & conference & health \\ \midrule
Diagnosing whether a person has the flu & conference & health \\ \midrule
Explaining the diagnosis and treatment options for the flu to a patient  & occupation & health \\ \midrule
Explaining the diagnosis and treatment options of cancer to a patient  & occupation & health \\ \midrule
Explaining the diagnosis and treatment options of depression to a patient  & occupation & health \\ \midrule
Helping stroke patients with physical rehabilitation, by guiding or assisting with exercise motions when needed (similar to what a physical therapist might do as part of their job) & conference & health \\ \midrule
Monitoring your health and alerting when you should go to the doctor & media & health \\ \midrule
Providing and coordinating patient care in a health facility, similar to a small part of what a Registered Nurse might do. & occupation & health \\ \midrule
Assisting an elderly person with showering or bathing & conference & living \\ \midrule
Brushing your teeth & life & living \\ \midrule
Buying groceries & life & living \\ \midrule
Cleaning up toxic waste, e.g., after a chemical spill & conference & living \\ \midrule
Cleaning your house & life & living \\ \midrule
Cooking dinner & life & living \\ \midrule
Deciding on an outfit for you to wear & conference & living \\ \midrule
Describing images or scenes for visually impaired people & conference & living  \\ \midrule
Editing an internet forum comment before you post it (e.g., for maximum popularity) & life & living \\ \midrule
Filling out and submitting your federal tax return paperwork & life & living \\ \midrule
Managing your personal finances/investments (similar to what a financial advisor might do) & life & living \\ \midrule
Monitoring a person's driving and intervening when they're distracted/in danger of making a mistake (e.g., emergency braking) & conference & living \\ \midrule
Tracking important moments and information and creating memory aids for elderly people & conference & living \\ \midrule
Translating an article you'd like to read from a foreign language to English & conference & living \\ \midrule
Asking a person out on a date & life & living, social \\ \midrule
Assisting your elderly parent & life & living, social \\ \midrule
Babysitting your child & life & living, social \\ \midrule
Breaking up with your romantic partner & life & living, social \\ \midrule
Finding people who might like to meet for a date & life & living, social \\ \midrule
Helping elderly individuals to increase their mobility by guiding them through crowded public spaces (e.g., walking to the grocery store) & conference & living, social \\ \midrule
Identifying the social relationship between two people (e.g., are they friends, a couple, strangers, siblings) & conference & living, social \\ \midrule
Picking out and buying a birthday present for an acquaintance & living & living, social \\ \midrule
Predicting the sexual orientation of a person & media & living, social \\ \midrule
Reading bedtime stories to your child & media & living, social \\ \midrule
Telling a joke & life & living, social \\ \midrule
Thinking of conversation topics while hanging out with friends & life & living, social \\ \midrule
Walking your dog & life & living, social \\ \midrule
Writing a birthday card to your mother & life & living, social \\

    \bottomrule
    \caption{The set of 100 tasks presented in our surveys.}
    \label{table:task_table}
\end{longtable}

\section{Survey Administration}
We advertise the survey on Amazon Mechanical Turk as a HIT (Human Intelligence Task) to workers who meet our quality screening guidelines.
Specifically, participants must have completed 200 HITs with at least a 99\% acceptance rate and must be from the United States.
Before accepting the HIT, participants are presented with the IRB-approved informed consent information, including the compensation amount (\$0.80) and a brief description of the survey and its purpose.
Upon providing informed consent and accepting the HIT, participants are presented with our survey.
Participants are only permitted to accept our HIT one time.

Upon accepting the HIT, participants are first shown the survey instructions, then the demographic questions.
Next, participants are presented with one randomly-selected task.
Participants evaluate the task under each component in our framework according to a five-point Likert scale.
Two attention questions are mixed in to this section.
Finally, participants choose the degree of delegation they would prefer for the task.
The questions are presented in a fixed order (not randomized).
Subjects are paid \$0.80 upon completing the survey and passing the checks attention check questions; otherwise the data is discarded.
Note that participants are not told which factor each question measures beyond the question text itself, and can choose the degree of delegation independently of our framework.
The HIT takes approximately 5 minutes to complete.
The full survey text is given in the next section.

\section{Survey Questions \& Demographics}
\subsection{Survey Instructions}
Note: You may only complete ONE HIT. Please do not queue hits or you will slow down study completion and delay payment.

We are conducting an academic survey about people’s attitudes towards “delegating” different kinds of tasks to an AI (artificial intelligence) versus to a person. You will:

Provide basic demographic information
Offer your opinion on properties of a task (e.g., mowing a lawn) in the form of agree/disagree statements
Choose the best way to divide control of the task between an AI and a person.

We expect the survey to take approximately 5-10 minutes, and you will be compensated \$0.80 upon submission (Expect approval within 1-2 days; Please note, your submission may not be approved if the attention questions are not answered correctly).

\subsection{Demographic Questions}

The following questions will help us to understand the study population and representativeness.

\textbf{1. What best describes your gender?}
\begin{itemize}
    \item Male
    \item Female
    \item Prefer to self-describe:
    \item Prefer not to say
\end{itemize}

\textbf{2. What is your age?}
\begin{itemize}
    \item 18-25
    \item 26-35
    \item 36-45
    \item 46-55
    \item 56-65
    \item 66-75
    \item 76 or older
    \item Prefer not to say
\end{itemize}

\textbf{3. How would you rate your level of computer proficiency?}
\begin{itemize}
    \item Far above average
    \item Slightly above average
    \item Average
    \item Slightly below average
    \item Far below average
    \item Prefer not to say
\end{itemize}

\textbf{4. What is the highest degree or level of school you have completed? (If currently enrolled, highest degree received)}
\begin{itemize}
    \item Some high school, no diploma, and below
    \item High school graduate, diploma or equivalent (for example: GED)
    \item Some college credit, no degree
    \item Trade/technical/vocational training
    \item Bachelor's degree, and above
    \item Prefer not to say
\end{itemize}

\subsection{Personal Survey Questions}
The following questions are the primary focus of this study.

\textbf{Important:} We will display a task. When answering the following questions, please carefully consider the task and your beliefs about an AI (artificial intelligence) agent performing the task, versus you personally performing the task.

(What is an AI agent? You can think of it as a computer, machine, robot, or some other form of automation.)

\textbf{Here is the task:}
$<$Task$>$

\textbf{5. For each of the following statements, please indicate the extent to which you agree or disagree for the above task:}
\begin{enumerate}
    \item This task requires social skills to complete.
    \item This task requires creativity to complete.
    \item This task requires a great deal of time or effort to complete.
    \item Attention check, please choose 'Strongly Agree' for this one.
    \item It takes significant training or expertise to be qualified for this task.
    \item I am confident in my own abilities to complete this task.
    \item In the case of mistakes or failure on this task, someone needs to be held accountable.
    \item A complex or unpredictable environment/situation is likely to cause this task to fail.
    \item Failure would result in a substantial negative impact on my life or the lives of others.
    \item I would feel motivated to perform this task, even without needing to; for example, it is fun, interesting, or meaningful to me.
    \item I am interested in learning how to master this task, not just in completing the task.
    \item I consider this task especially valuable or important; I would feel committed to completing this task because of the value it adds to my life or the lives of others.
    \item Attention check, please choose 'Strongly Disagree' for this one.
    \item I trust the AI agent's ability to reliably complete the task.
    \item Understanding the reasons behind the AI agent's actions is important for me to trust the AI agent on this task (e.g., explanations are necessary).
    \item I trust the AI agent's actions to protect my interests and align with my values for this task.
\end{enumerate}

\textbf{6. If you were to do the given (above) task, what level of AI/machine assistance would you prefer?}
\begin{enumerate}
    \item \textbf{Full AI automation:} decisions and actions are made automatically by the AI once the task is assigned; you do nothing.
    \item \textbf{The AI leads and the human assists:} the AI performs the task, but asks you for suggestions/confirmation when appropriate.
    \item \textbf{The human leads and the AI assists:} you do the task mostly on your own, but the AI offers recommendations or help when appropriate (e.g., you get stuck or AI sees possible mistakes).
    \item \textbf{No AI assistance:} you do the task completely on your own.
\end{enumerate}

\subsection{Expert Survey Questions}
The following questions are the primary focus of this study.

\textbf{Important:} We will display a task. When answering the following questions, please carefully consider the task and your beliefs about an AI (artificial intelligence) agent performing the task, versus a qualified human person performing the task.

(What is an AI agent? You can think of it as a computer, machine, robot, or some other form of automation.)

\textbf{Here is the task:}
$<$Task$>$

5. \textbf{For each of the following statements, please indicate the extent to which you agree or disagree for the above task:}
\begin{enumerate}
    \item This task requires social skills to complete.
    \item This task requires creativity to complete.
    \item This task requires a great deal of time or effort to complete.
    \item Attention check, please choose 'Strongly Agree' for this one.
    \item It takes significant training or expertise to be qualified for this task.
    \item I am confident in a qualified person's ability to complete this task.
    \item In the case of mistakes or failure on this task, someone needs to be held accountable.
    \item A complex or unpredictable environment/situation is likely to cause this task to fail.
    \item Failure would result in a substantial negative impact on my life or the lives of others.
    \item Attention check, please choose 'Strongly Disagree' for this one.
    \item I trust the AI agent's ability to reliably complete the task.
    \item Understanding the reasons behind the AI agent's actions is important for me to trust the AI agent on this task (e.g., explanations are necessary).
    \item I trust the AI agent's actions to protect my interests and align with my values for this task.
\end{enumerate}

6. \textbf{If you were to ask someone to complete the given (above) task, what level of AI/machine assistance would you prefer?}
\begin{enumerate}
    \item \textbf{Full AI automation:} decisions and actions are made automatically by the AI once the task is assigned; no human involvement.
    \item \textbf{The AI leads and the human assists:} the AI performs the task, but asks the person for suggestions/confirmation when appropriate.
    \item \textbf{The human leads and the AI assists:} the person does the task mostly on their own, but the AI offers recommendations or help when appropriate (e.g., human gets stuck or AI sees possible mistakes).
    \item \textbf{No AI assistance:} the person does the task completely on their own.
\end{enumerate}

\para{Demographics.}
We have two surveys (delegating to experts vs AI, or the subject personally vs AI), 100 tasks per survey, and 4 or 5 responses per task. 
Of the 992 subjects (495 in the \personalsurvey and 497 in the \expertsurvey), 525 identified as male, 463 as female, 2 as non-binary, and 2 preferred not to indicate. 136 were aged 18-25, 421 aged 26-35, 223 aged 36-45, 118 aged 46-55, 78 aged 56-65, 13 aged 66-75, 1 aged 76+, and 2 preferred not to indicate.

\end{document}